\documentclass[Journal,NoLineNumbers,InsideFigs]{ascelike-new}
\usepackage[utf8]{inputenc}
\usepackage[T1]{fontenc}
\usepackage{lmodern}
\usepackage{graphicx}
\usepackage[style=base,figurename=Fig.,labelfont=bf,labelsep=period]{caption}
\usepackage{subcaption}
\usepackage{amsmath}
\usepackage{booktabs}
\usepackage{siunitx} 
\usepackage{hologo}
\usepackage[utf8]{inputenc}
\usepackage{float}
\usepackage{algorithm}
\usepackage{algpseudocode}
\usepackage{booktabs}
\usepackage{tabularx}
\usepackage{comment}
\usepackage{csquotes} 
\usepackage{xcolor}
\usepackage{amsmath,amssymb,amsfonts}
\usepackage{graphicx}
\usepackage{textcomp}
\usepackage{adjustbox}
\usepackage{placeins}
\usepackage{newtxtext,newtxmath}
\usepackage{float}

\usepackage[colorlinks=true,citecolor=red,linkcolor=black]{hyperref}
\NameTag{Akhavian, \today}

\begin{document}

\title{Building Information Models to Robot-Ready Site Digital Twins (BIM2RDT): An Agentic AI Safety-First Framework}

\author[1]{Reza Akhavian$^{\ast}$}
\author[1,2]{Mani Amani}
\author[1,3]{Johannes Mootz}
\author[4]{Robert Ashe}
\author[4]{Behrad Beheshti}

\affil[1]{Department of Civil, Construction, and Environmental Engineering, San Diego State University, San Diego, CA, United States.}
\affil[2]{Department of Electrical and Computer Engineering, University of California, San Diego, San Diego, CA, United States}
\affil[3]{Department of Mechanical and Aerospace Engineering, University of California, San Diego, San Diego, CA, United States}
\affil[4]{Department of Computer Science, San Diego State University, San Diego, CA, United States\par
*Corresponding author. Email: rakhavian@sdsu.edu}

\maketitle

\section{Abstract}

The adoption of cyber-physical systems and jobsite intelligence that connects design models, real-time site sensing, and autonomous field operations can dramatically enhance digital management in the Architecture, Engineering, and Construction (AEC) industry. This paper introduces BIM2RDT (Building Information Models to Robot-Ready Site Digital Twins), an agentic artificial intelligence (AI) framework designed to transform static Building Information Modeling (BIM) into dynamic, robot-ready digital twins (DTs) that prioritize safety during construction execution. The framework bridges the gap between pre-existing BIM data and real-time site conditions by integrating three key data streams: geometric and semantic information from BIM models, real-time activity data from IoT sensor networks, and visual-spatial data collected by quadruped robots during site traversal. The methodology introduces Semantic-Gravity ICP (SG-ICP), a novel point cloud registration algorithm that leverages large language model (LLM) reasoning. Unlike traditional methods, SG-ICP utilizes an LLM to infer object-specific, physically-plausible orientation priors based on BIM semantics, significantly improving alignment accuracy by avoiding convergence on local minima. This creates an intelligent feedback loop where robot-collected data updates the DT, which in turn optimizes paths for subsequent missions. The framework employs YOLOE open-vocabulary object detection and Shi-Tomasi corner detection to identify and track construction elements while using BIM geometry as robust \textit{a priori} maps. The framework also integrates real-time Hand-Arm Vibration (HAV) monitoring, mapping sensor-detected safety events to the digital twin using IFC standards for proactive intervention. Major findings from experiments demonstrate SG-ICP's superiority over standard ICP, achieving RMSE reductions of 64.3\%–88.3\% in alignment across varied scenarios with occluded or sparse features, ensuring physically plausible orientations. HAV integration triggers real-time warnings and tasks upon exceeding exposure limits, enhancing compliance with such standards as ISO 5349-1.

\section{Practical Applications}

Construction sites are becoming increasingly complex with the introduction of new technologies such as reality capture equipment and robots, requiring better tools to streamline adoption, avoid tool sprawl, and ensure worker safety. This research introduces a system that combines robots, smart sensors, and building information modeling (BIM) data to create a “digital twin”: an up-to-date virtual copy of a construction site's geometries and safety information. The system uses quadruped robots equipped with cameras and sensors to autonomously walk through construction sites, automatically detecting and tracking objects like equipment, materials, and temporary structures. Unlike traditional approaches that start from scratch, this method leverages existing BIM data as a foundation, making the robots more accurate and efficient at understanding their surroundings. Besides geometric site updates, safety information is also presented in the updated digital twin. Furthermore, the system integrates with wearable sensors to monitor worker safety, such as detecting excessive exposure to tool vibrations, and automatically flags potential hazards for safety managers. This helps prevent long-term health issues like hand-arm vibration syndrome while maintaining detailed safety records. This technology helps bridge the gap between office plans and field reality, paving the way for safer, more efficient, and more automated construction sites.

\section{Introduction}

The Architecture, Engineering, and Construction (AEC) industry is in the midst of a profound digital transformation. Historically characterized as a late adopter of digital technology, the sector has traditionally relied heavily on the experiential knowledge of its professionals. However, growing pressures from economic uncertainties due to factors such as low productivity, persistent labor shortages, and the increasing complexity of modern projects are compelling companies to rapidly adopt digital construction management strategies to optimize resources and workflows. This digital revolution is fundamentally redefining the entire project lifecycle, from initial planning and design through to long-term maintenance and operations. The current industry environment demands data-driven workflows and modern project management software that empower teams to make informed, real-time decisions. Market projections indicate that adoption of technologies such as the Digital Twins (DTs) in the construction market will reach \$155.01 billion by 2030, with safety applications and worker monitoring driving significant growth \cite{DigitalTwinConstruction2030}.

In response to these challenges, the concept of Cyber-Physical Systems (CPS) has emerged as a transformative paradigm for the construction industry. A CPS integrates computation, networking, and physical processes, creating a feedback loop where physical systems are monitored and controlled by computer-based algorithms \cite{LEE201518}. This paradigm is built upon four core technological pillars that are increasingly converging to create a cohesive, intelligent construction management ecosystem. The first pillar is Building Information Modeling (BIM), which serves as the foundational digital blueprint, providing a data-rich, object-oriented model of the planned asset. The second is the DTs, which represent the dynamic evolution of BIM, transforming the static model into a live, virtual replica of the physical site through real-time data integration. The third pillar is Robotics, which provides the autonomous physical layer for data acquisition and task execution, bridging the gap between the digital and physical worlds. Finally, Agentic Artificial Intelligence (AI) acts as the cognitive engine of the system, processing vast amounts of data, making autonomous decisions, and coordinating actions to achieve project goals. Together, these pillars form the basis of a next-generation approach to construction management that moves beyond manual processes and reactive decision-making \cite{TUHAISE2023104931}.

\begin{figure}
  \centering
  \includegraphics[width=\columnwidth]{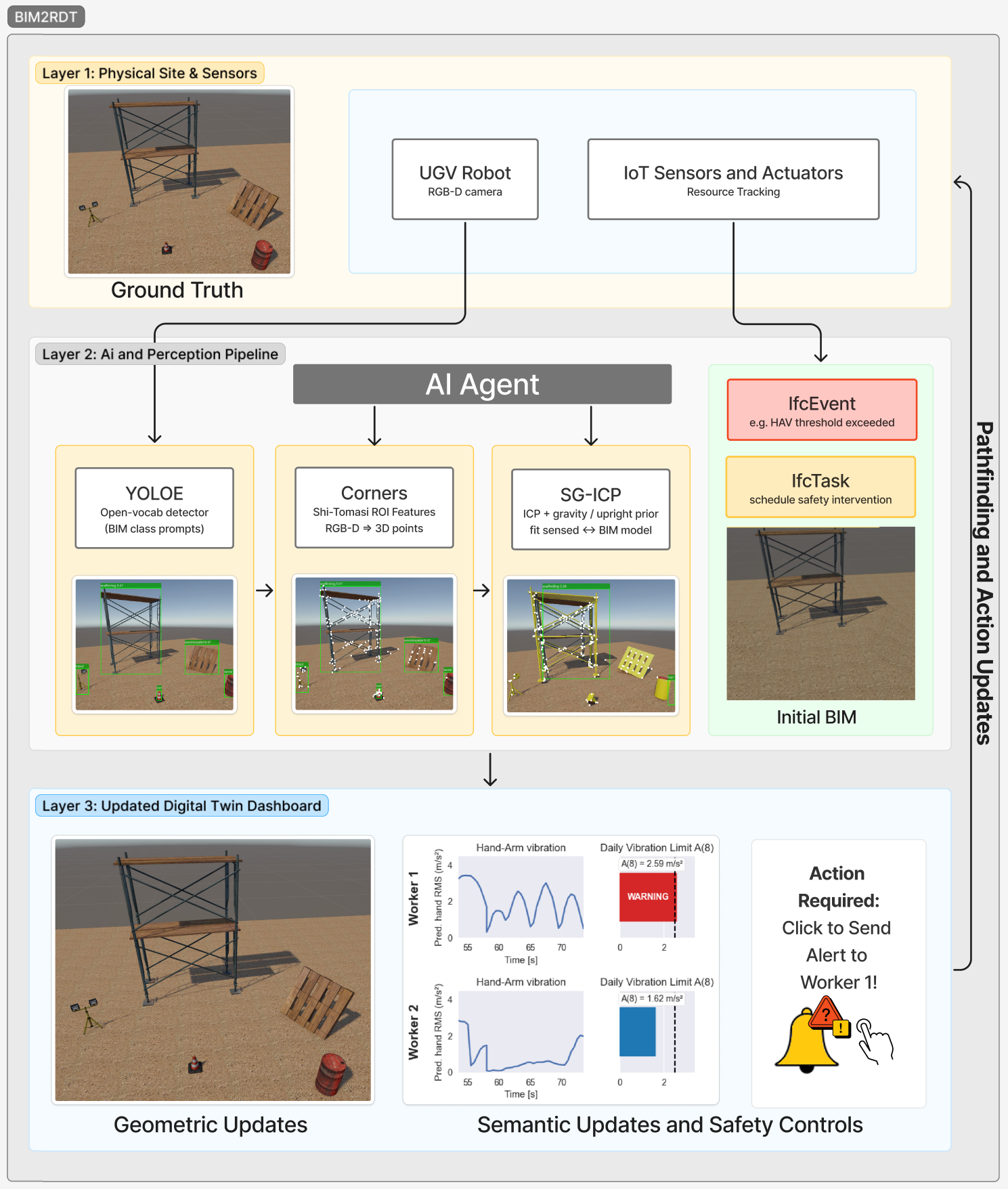}
  \caption{Overview of the BIM2RDT framework.}
  \label{fig:OverallFigure}
\end{figure}

This paper presents BIM2RDT, \underline{\textbf{B}}uilding \underline{\textbf{I}}nformation \underline{\textbf{M}}odels \underline{\textbf{to}} \underline{\textbf{R}}obot-\underline{\textbf{R}}eady Site \underline{\textbf{D}}igital \underline{\textbf{T}}wins, where the central '2' embodies both our BIM transformational approach and the dual 'R' requirements for robot readiness. BIM2RDT is an agentic AI framework that transforms static BIM models into dynamic digital twins with virtual components for robot path planning, activity recognition, and safety monitoring and compliance during construction execution. This addresses a critical gap in current construction robotics: the lack of integrated systems that simultaneously leverage pre-existing BIM data, real-time sensor streams, and autonomous decision-making for safe and efficient site navigation. The proposed methodology integrates three key data streams: (1) geometric and semantic information from BIM models as the foundational spatial and functional representation, (2) real-time activity data from IoT sensor networks monitoring site resources such as workers and equipment, and (3) visual and spatial data collected by quadruped robots during site traversal. This multi-modal data fusion and semantic understanding using an agentic AI engine enables continuous digital twin updates that reflect actual site conditions, including temporary obstacles, active work zones, activity types, and safety hazards not captured in static BIM models. Figure \ref{fig:OverallFigure} shows an overview of the developed framework.

Unlike traditional SLAM (Simultaneous Localization and Mapping) approaches that operate in completely unknown environments, BIM2RDT leverages existing BIM geometry as a robust \textit{a priori} map. The framework creates an intelligent feedback loop where robot-collected data validates and updates the digital twin, which in turn generates optimized path plans for subsequent robot missions. This addresses three fundamental SLAM challenges in construction environments: (1) drift accumulation in large-scale sites through BIM-anchored localization, (2) dynamic environment changes through continuous model updates, and (3) GPS-denied indoor navigation through semantic BIM features as landmarks. Finally, the safety-first design prioritizes worker protection by integrating wearable sensor data to establish dynamic exclusion zones and using agentic AI to make real-time navigation decisions that minimize human-robot interaction risks while maintaining operational efficiency.

\section{Literature Review}

\subsection{BIM-to-Digital Twin Transformation}

The evolution from Building Information Models (BIM) to digital twins (DTs) in construction represents a paradigm shift toward data-driven, real-time management of built environments. The defining characteristic that distinguishes a true DT from an advanced BIM model is this bi-directional, real-time or near-real-time data connection \cite{jiang2024digital}. This cyber-physical integration establishes a continuous feedback loop: the physical asset is instrumented with sensors that feed data to the digital model, and the digital model, in turn, can be used to analyze, simulate, and control the physical asset.  This concept is still nascent in construction but growing rapidly as part of the Construction 4.0 paradigm \cite{TUHAISE2023104931}. A 2021 systematic review analyzed 123 papers, categorizing DT applications into a five-level ladder based on building lifecycle stages: Level 1 (basic BIM for visualization), Level 2 (BIM-supported simulations), Level 3 (BIM with IoT for monitoring), Level 4 (BIM with AI for predictions), and Level 5 (ideal DTs with automated control) \cite{deng2021bim}. Key findings indicate most studies remain at Levels 2–3, with IoT integration enhancing energy performance and indoor environment monitoring, but gaps in real-time simulations and data interoperability persist. Therefore, the fundamental limitation of BIM is its static nature: a BIM model primarily captures the as-designed or as-planned intent of a project, serving as a digital blueprint created before or during construction. It does not inherently reflect the dynamic, ever-changing reality of the physical construction site.

The transformation of a static BIM into a dynamic DT is made possible by a suite of enabling technologies that bridge the physical and digital realms. The cornerstone of this transition is the integration of the IoT, which encompasses a network of sensors, devices, and communication protocols that allow for the collection and exchange of real-time data. This integration of IoT and BIM is widely considered to have paved the way for the emergence of the DT concept in the built environment \cite{sepanosian2025iot}. Sensor networks are deployed on the construction site to capture a continuous stream of data about physical assets, processes, and environmental conditions. Specific technologies commonly used include Radio-Frequency Identification (RFID) and Ultra-Wideband (UWB) for tracking the location and identity of materials, equipment \cite{akhavian2013simulation}, and personnel \cite{akhavian2015wearable}. Embedded sensors can monitor structural health, and environmental sensors can provide data on temperature, humidity, and other conditions that may affect construction activities \cite{shahnavaz2022automated}. This multi-sourced data is transmitted via communication gateways to a cloud-based platform where it is processed and integrated with the virtual model, ensuring the DT remains a synchronized, high-fidelity representation of the physical site.

Once established, a construction DT offers a wide range of functional capabilities that significantly enhance project management and operational efficiency. Its primary function is to provide real-time monitoring and increased transparency of information, allowing stakeholders to have a deep and current understanding of project progress and asset conditions. This capability reduces the need for costly and time-consuming site visits and provides a central source of information that boosts productivity and collaboration \cite{pal2023automated}. Beyond simple monitoring, DTs enable advanced simulation and analysis. By modeling various scenarios, facility managers can predict maintenance requirements, optimize space utilization, and monitor energy consumption in real-time, reducing downtime and extending the lifespan of building systems \cite{akhavian2014construction}. This predictive power allows for timely, data-driven decisions to be made in a proactive manner.

\subsection{AI and Autonomous Agents in Construction}

Within the broad field of AI, the concept of AI agents represents a significant step beyond simple automation or traditional predictive analytics. Unlike automation tools that follow rigid, pre-programmed instructions, or analytical models that simply provide outputs for human interpretation, AI agents are software systems designed to perceive their digital or physical environment, reason about that information, and act autonomously to achieve specified goals. A key characteristic of an AI agent is its ability to operate without constant human input, interpreting data, assessing progress against a goal, and adjusting its actions in real-time to adapt to changing conditions \cite{hosseini2025role}. This capacity for autonomous, goal-directed action makes them particularly well-suited for managing the unpredictable and dynamic nature of construction projects, where designs shift and resource availability fluctuates hourly.

In the construction context, AI agents are being developed to handle a wide range of tasks, moving beyond mere analysis to active execution. Their applications include automating complex administrative and logistical workflows, optimizing the allocation of resources like labor and equipment, and enhancing collaborative decision-making. For example, an AI agent can be integrated into a company's existing enterprise systems, such as project management platforms or ERP software, to act as a digital intermediary \cite{jamithireddy2025automation}. In this role, it can autonomously monitor subcontractor compliance, track invoice approvals, or flag budget variances based on predefined rules, initiating the next step in a workflow once certain criteria are met. Monitoring building operations is another field where agentic AI can be effectively deployed \cite{both2023automated}. Although the dynamic and complex nature of construction execution stands to gain significantly from the integration of autonomous AI agents, this domain remains relatively nascent in the academic literature. Consequently, the development of intelligent agents capable of perception, planning, and action in unstructured construction environments represents a critical and underexplored research frontier \cite{lee2025agentic}.

\subsection{Robotics in Site Monitoring and Data Collection}

The continuous and accurate acquisition of as-built data is the lifeblood of a Digital Twin. While fixed sensors provide valuable data streams, mobile robotic platforms have emerged as a primary method for comprehensive and autonomous site scanning, addressing the limitations of manual data collection, which is often time-consuming, error-prone, and unsafe  \cite{halder2022real}. In construction, various Unmanned Ground Vehicles (UGVs) have been developed for ground-level inspection tasks. These platforms, which range from wheeled robots to wall-climbing and cable-crawling systems, offer a safer and more efficient alternative to manual inspection, especially in hazardous or difficult-to-access areas \cite{mantha2018robotic}. Among different UGV types, quadruped robots offer unprecedented agility and the ability to dynamically navigate cluttered spaces, self-right after a fall, and traverse complex topographies. This superior mobility makes them exceptionally well-suited for conducting consistent, repeatable, and autonomous data collection missions inside buildings during various stages of construction, an environment where other robotic platforms may struggle or fail entirely \cite{halder2022accuracy}. Additionally, their ability to operate without constant human intervention, autonomously navigating pre-planned routes while avoiding new obstacles, makes them a reliable and scalable tool for routine site monitoring \cite{ashe2025modular}.

Despite their promise, the deployment of autonomous robots on construction sites faces several practical and technical challenges. Robust and reliable navigation in highly dynamic and constantly changing environments remains a significant hurdle \cite{hu2025semantic}. While quadrupeds are agile, they must still be able to accurately localize themselves, often in GPS-denied indoor environments, and dynamically replan paths to avoid unforeseen obstacles like temporary materials, equipment, or personnel. The fusion and registration of data collected from multiple scans and different sensor modalities (e.g., LiDAR, cameras) into a single, coherent model is another complex technical problem \cite{chen2019deep,amani2025safe}. Moreover, achieving full autonomy is often not feasible or desirable. Instead, research is increasingly focused on developing effective human-robot teaming and collaboration strategies, where robots act as assistants to human inspectors, augmenting their capabilities rather than replacing them entirely \cite{amani2024intelligent}. This partnership approach leverages the strengths of both humans (e.g., expert judgment, complex problem-solving) and robots (e.g., tireless and precise data collection) to create a more effective inspection workflow \cite{yu2024cloud}.

\subsection{Geometric Registration with ICP}

The accurate alignment of 3D point clouds is a cornerstone of robotics, computer vision, and digital twinning for the built environment. The Iterative Closest Point (ICP) algorithm remains the foundational method for this task, but its limitations have inspired decades of innovation. The standard ICP algorithm iteratively refines the rigid transformation between two sets of point clouds by minimizing the sum of squared distances between corresponding points \cite{besl1992method}. Despite its widespread adoption, ICP's performance is highly sensitive to the initial alignment and is famously susceptible to converging to local minima, especially in scenes with sparse features, repetitive geometry, or poor initial overlap \cite{pomerleau2015review}. A significant leap in ICP's robustness came from constraining the 6-DoF (Degrees of Freedom) search space of the Special Euclidean group SE(3). By incorporating external information, the non-convex optimization problem can be further simplified depending on the implementation context, reducing the search space to a 4-DoF problem (x, y, z, yaw), and drastically decreasing the likelihood of falling into local minima and accelerating convergence \cite{koide2021voxelized}. Suppose the objects are known to be restricted to 4 DoF (e.g., concrete walls, UGV, pillars). In that case, we can restrict the DoF to conduct optimization in SE(2), which would reduce the possibility for convergence on local minima. At the same time, hard constraints such as reducing DoF may not be a feasible assumption given different contexts. Instead, other works have introduced soft constraints to promote specific orientation initialization to aid ICP \cite{iMUICP,ding2021globally}. 

In construction, robust point cloud registration (often referred to as "scan-to-BIM") is critical for as-built verification, progress monitoring, and digital twin maintenance. Researchers have extensively used ICP and its variants to align laser scans of a job site with the as-designed BIM model to detect deviations \cite{abreu2023procedural}.The structured nature of building environments, with prevalent planar surfaces, makes point-to-plane ICP particularly effective. Global BIM-point cloud registration minimizes deviations using ICP, evaluating alignments for construction quality control \cite{zhang2024global}. Lidar inertial odometry (LIO) and BIM have been used together for robust robot localization, using ICP-like methods to map and navigate in building environments \cite{stuhrenberg2025lio}. Semantic-aided ICP in global LiDAR registration uses BIM semantics via Pose Hough Transform, enhancing robot autonomy in construction \cite{qiao2025speak}. Online as-built BIM updates for robotic monitoring solve SLAM with ICP, enabling real-time model refinements \cite{spinner2024online}. However, in all prior work, a fixed upright prior based on external (e.g., IMU) data have been used. In contrast, the proposed method uses an LLM's "common sense" physical reasoning to create a dynamic, object-specific upright prior. The LLM's ability to infer that a heavy "crate" is more likely to be upright than a "traffic cone" allows it to assign a physically-informed, continuous bias to the ICP algorithm. Furthermore, this approach does not require a visually trained deep learning model for object classification; instead, it leverages the vast world knowledge embedded in the LLM to parameterize the geometric optimization.

\section{Methodology}
\subsection{Object Detection}
We employ the YOLOE open-vocabulary object detector \cite{wang2025yoloe}. The selection of YOLOE is deliberate and critical to the BIM2RDT framework. Unlike traditional closed-set object detectors, which are limited to a predefined set of categories, open-vocabulary models can identify objects based on arbitrary textual descriptions. This capability is particularly suitable for construction sites, which are populated with a broad assortment and often variable set of object labels, from standard BIM components (e.g., columns, beams) to temporary materials (e.g., pallets, scaffolding) and unplanned obstacles. YOLOE allows our system to be highly adaptable, as it can be dynamically prompted with the specific BIM families relevant to the robot's current operational area without requiring model retraining, a significant advantage over alternatives.

Semantic information from the BIM or user-specified labels initializes a textual prompt that biases classification toward the target BIM families. The process begins by translating the semantic information from the BIM into a concise textual prompt for the detector. BIM family names, which can be verbose or contain non-descriptive metadata (e.g., \texttt{"Steel\_I-Beam\_W12x26\_A992"}), are processed using an LLM to extract the core object class. For example, the aforementioned name is simplified to ``steel I-beam''. Similarly, \texttt{"Traffic\_Cone\_Orange\_70cm"} becomes ``traffic cone''. The prompt concatenates BIM family names, such as ``Concrete Pillar'' or ``Safety Cone'', with contextual descriptors like ``construction site object'' to bias classification. As an example, for a BIM family ``Wall'' the prompt might be: ``a concrete wall in a building site''.

This list of simplified class names, derived from the BIM elements expected in the scene, forms the textual prompt. For a given region, the prompt might be a list such as: ["steel I-beam", "traffic cone", "wooden crate", "steel barrel"]. This context-specific prompt biases the classification towards the target BIM families, improving detection accuracy and reducing false positives. The detector outputs bounding boxes with associated class names, which we then map one-to-one to the corresponding BIM families. These mappings define regions of interest (ROI) that support downstream feature selection and 3D registration analyses. In dynamic sites, this semantic biasing reduces false positives from clutter, ensuring ROIs align with BIM elements for accurate deviation detection.

\subsection{Corner Detector}
In order to detect features within the 2D ROI, the shape geometry is required. We opt to use Shi-Tomasi corner detection to detect these features due to the algorithm's superior stability in tracking under affine transformations, which is beneficial for construction scenes with perspective distortions from robot viewpoints \cite{shitomassi}. It is important to note that any corner detector algorithm can be used in this framework. The goal of the corner detector is to identify coordinates that exist on the object, and this set of 3D coordinates would give us plausible points that exist on the object geometry that can be used for point cloud alignment. Given a grayscale image $I$, compute spatial gradients $I_x,I_y$.
For each pixel $x$ in the ROI, form the second-moment matrix over a window $W(x)$
\[
M(x)=\sum_{u\in W(x)}
\begin{bmatrix}
I_x(u)^2 & I_x(u)I_y(u)\\
I_x(u)I_y(u) & I_y(u)^2
\end{bmatrix}.
\]
Let $\lambda_1(x)\ge \lambda_2(x)$ be the eigenvalues of $M(x)$. The Shi--Tomasi score is
\[
S(x)=\min\{\lambda_1(x),\lambda_2(x)\}.
\]
Accept $x$ if $S(x)\ge\tau$ (cornerness threshold), then apply non-maximum suppression in a radius $r$, and keep the top $K$ responses. The threshold $\tau$ is adaptively set based on image noise levels, typically $0.01-0.05$, with $r=3$ pixels and $K=600$ for construction ROIs to balance density and computational efficiency. The Shi-Tomasi algorithm is a cornerstone in feature detection, known for its robustness and availability across OpenCV implementations. These detected corners within BIM-mapped ROIs provide robust keypoints for ICP correspondence. Since we are using a depth camera to provide perception to the robot, we can use the identified pixel \((x,y)\) to query the \(z\) value so we get a 3D representation of the feature values that are detected on the 2D image. Given set of points \(A = \{(x_i, y_i, z_i)\}_{i=1}^N\) where \(A \subset \mathbb{R}^3\), we can then conduct ICP with a root set of points \(B = \{(x_i, y_i, z_i)\}_{i=1}^N\) where \(B \subset \mathbb{R}^3\), which becomes our target points sampled from the BIM.

\subsection{ICP}
In BIM2RDT, the ICP algorithm aligns robot-captured point clouds (e.g., from LiDAR mounted on quadrupeds) with BIM-derived models, enabling deviation detection for progress monitoring \cite{bosche2010automated}. The goal of using ICP is to minimize Wahba's problem \cite{WahbasProblem}, given by:
\[
\min_{R\in \mathrm{SO}(3),\; t\in \mathbb{R}^3}
\sum_{i=1}^{N} w_i\,\left\lVert \mathbf{p}_i - \bigl(R\,\mathbf{q}_i + t\bigr) \right\rVert^2
\]
While closed-form solvers such as the Kabsch algorithm can recover \((R,t)\) given a set of \emph{known} point correspondences~\cite{Kabsch:a12999},
they require the data association to be specified \textit{a priori}. However, the association is often not known in practice. To address this, ICP was introduced, which initializes data classifications given a defined heuristic. Concretely : ICP alternates
between (i) estimating correspondences (nearest-neighbor matching under the current pose) and (ii) solving an optimization step for \((R,t)\),
until convergence or until a maximum number of iterations has been reached. This joint treatment allows ICP to handle unknown data association in practice.
\par
However, ICP is known to converge to local minima. To mitigate this issue, many methods have been proposed. One of these methods is to reduce the degrees of freedom by introducing gravity as an additional source of information to improve point cloud alignment. Suboptimal initial conditions narrow ICP’s basin of attraction, increasing local-minimum failures on construction geometries. This can be detrimental to BIM updates and potentially introduce discrepancies. This problem is addressed in BIM2RDT via semantic priors, and is explained in the following subsection.
\subsection{Semantic-Gravity ICP}
Consider the regularized optimization problem:
\[
\min_{R\in\mathrm{SO}(3),\,t\in\mathbb{R}^3}
\;\frac12\sum_{i=1}^N w_i\,\|\,b_i - (R r_i + t)\|_2^2
\;+\; \frac{\gamma}{2}\,\|\,g_m - R g_0\|_2^2 .
\]
where $r_i \in \mathbb{R}^3$ are model sample points with normals, $b_i \in \mathbb{R}^3$ are corresponding scene points, $w_i \ge 0$ are robust weights, $(R,t) \in \mathrm{SO}(3)\times \mathbb{R}^3$ is the unknown rigid transform, $g_0$ is the canonical upright vector in model coordinates (e.g.\ $e_z$), $g_m$ is the measured gravity vector in world coordinates, and $\gamma \ge 0$ is the gravity weight, which balances data fidelity and upright bias; in practice, we tune it via cross-validation on simulated construction scenes.
\subsubsection{Gravity Alignment}
The data term
\[
E_{\mathrm{ICP}}(R,t) = \tfrac{1}{2} \sum_i w_i \, \|\, b_i - (R r_i + t)\|_2^2
\]
is a standard weighted ICP.

The gravity bias term
\[
E_{\mathrm{grav}}(R) = \tfrac{\gamma}{2} \, \|\, g_m - R g_0 \|_2^2
\]
acts as a soft constraint, encouraging $R g_0$ to align with gravity, maintaining a more upright orientation.
Expanding the gravity penalty:
\[
E_{\mathrm{grav}}(R)
= \frac{\gamma}{2}\big(\,\|g_m\|_2^{2} + \|R g_0\|_2^{2} - 2\, g_m^{\top} R g_0 \big).
\]
Since $\|g_m\|=\|g_0\|=1$ in unit-norm form, this reduces to
\[
E_{\mathrm{grav}}(R) = \gamma\big(1 - g_m^{\top} R g_0\big).
\]
Thus, minimizing $E_{\mathrm{grav}}$ is equivalent to maximizing the alignment of $R g_0$ with $g_m$.

In Gauss–Newton or SVD updates, this bias is equivalent to adding one weighted correspondence $(g_0,\,g_m)$ with weight $\sqrt{\gamma}$:
\[
\{(r_i,b_i,\sqrt{w_i})\}_{i=1}^N \;\cup\; (g_0,\,g_m,\,\sqrt{\gamma}).
\]

\par
\subsubsection{Semantic Alignment}
We use BIM semantic labels to estimate each object's susceptibility to gravity-induced tipping. Heavier objects tend to have a stronger upright bias; if they appear tilted, routine site dynamics or workspace drift are less likely explanations. This extends prior gravity-constrained ICP (e.g., IMU-ICP \cite{iMUICP}) by making the bias object-adaptive via BIM semantics, reducing errors in tilted but stable objects like heavy equipment.
We define 
\[
\lambda_{new} = \gamma_f .\gamma_{initial}
\]
where \(\lambda_{initial}\) is the initialized value that can serve as a general tuning knob. 
Assigning weights manually is cumbersome, so we instead use an LLM to traverse all BIM families and assign \(\gamma_{\text{family}} \in [0,1]\), a prior for the object being upright. The LLM interprets family names, descriptions, and construction-site context and outputs the corresponding bias score.
 We use the OpenAI API to prompt GPT-4o-mini. Specifically, the LLM is given the following context: 
 \begin{quote}
     You return BIM alignment priors. For each label, output: upright bias $\in [0,1]$ (fraction of tilt corrected per ICP iteration). Only include labels given. No extra text. Take gravity into account, how likely is this to not be upright? If it is not likely we need a smaller value. JSON that matches the provided schema strictly.
 \end{quote}
A JSON file containing the name of all families is attached to the command. The LLM will use its contextual reasoning capabilities to infer the possibilities of being "upright" and returns a tailored coefficient.  
\begin{algorithm}
\caption{SG-ICP}
\label{alg:gc-icp}
\begin{algorithmic}[1]
\Require body $B$; loss $\mathcal{L}$, schedule $L$, default bias $\beta_{\mathrm{def}}$; gravity $\hat g$; params: \textit{restarts}, \textit{yawJitterDeg}, \textit{sampleCount}, \textit{maxIterations}, \textit{maxPairDistance}, \textit{normalGate}, \textit{trimFraction}, \textit{huberDelta}, $(\varepsilon_r,\varepsilon_t)$
\Ensure pose of $B$ aligned to scene $S$
\State $(\beta,\text{yawOnly}) \gets \textsc{LLMBiasManager}(\mathcal{L},L,\beta_{\mathrm{def}})$
\State $S \gets \textsc{AcquireScenePoints}()$ \quad \textbf{require } $|S|\ge 6$
\State $(V,N) \gets \textsc{SampleModelWithNormals}(B,\textit{sampleCount})$
\State $(p_0,q_0) \gets \textsc{SnapshotPose}(B)$; \textsc{PrepareMovementMode}$(B)$
\For{$a \in \textsc{YawRestarts}(\textit{restarts},\hat g,\textit{yawJitterDeg})$}
  \State \textsc{ResetPose}$(B,p_0,q_0)$; \textsc{ApplyYawJitter}$(B,a)$
  \For{$k=1..\textit{maxIterations}$}
    \State $(X,N_W) \gets \textsc{ModelToWorld}(B,V,N)$
    \State $P \gets \textsc{MatchPairs}(X,S,\textit{maxPairDistance},N_W,\textit{normalGate})$
    \State $P \gets \textsc{RelaxIfFew}(P,S,X)$; \quad $P \gets \textsc{TrimPairs}(P,\textit{trimFraction})$
    \State $w \gets \textsc{RobustWeights}(P,\textit{huberDelta})$
    \State $(R_\Delta,t_\Delta) \gets \textsc{SolveRigid}(P,w)$
    \State $(R_\Delta,t_\Delta) \gets \textsc{ApplyBias}(B,\hat g,\beta,P,w,R_\Delta,t_\Delta)$
    \State $R_\Delta \gets \textsc{ProjectYawIf}(R_\Delta,\hat g,\text{yawOnly})$
    \State \textsc{ApplyDelta}$(B,R_\Delta,t_\Delta)$
    \If{\textsc{Converged}$(P,R_\Delta,t_\Delta,\varepsilon_r,\varepsilon_t)$} \textbf{break} \EndIf
  \EndFor
  \State \textsc{ScoreAndKeepBest}$(B,S,\textit{maxPairDistance})$
\EndFor
\State \textsc{SetBestPose}$(B)$; \textsc{RefreshGizmos}$(B,V,N,S)$; \textsc{CleanupMovementMode}$(B)$
\end{algorithmic}
\end{algorithm}

\subsection{Safety-Aware Digital Twin Updates with Activity Integration}

Once the robot-captured point cloud is geometrically registered to the BIM using the Semantic-Gravity ICP (SG-ICP) method, the BIM2RDT framework extends these capabilities to incorporate real-time activity recognition for enhanced safety monitoring. Here, it semantically links sensor-detected activities to BIM entities using IFC standards. Specifically, it leverages \textit{IfcTask} and \textit{IfcEvent} (core entities in the IFC schema for representing identifiable units of work or activities) as a unifying interface, allowing sensor-derived activities (e.g., "hammer drilling with unsafe vibration levels") to be treated as compatible IFC constructs without requiring explicit temporal scheduling ties. 

The integration of dynamic site conditions beyond geometry is exemplified by the analysis of worker activities involving hazards such as excessive hand-arm vibrations (HAV) during tool operation.  We draw from prior work on wearable sensor-based HAV monitoring \cite{mootz2025advancing}, which demonstrated the feasibility of attaching wearables to positions beyond those specified in ISO 5349-1 \cite{ISO2001a-5349-1}. In this work, we expand upon the existing methodology and develop a subject-specific approach that enables more accurate mapping.

The acceleration signals are transmitted to a central processing unit, where the frequency-weighted RMS acceleration $a_{hv}$ is calculated. This data serves as the input for computing \textit{A(8)}, the 8-h energy-equivalent frequency-weighted vibration total value, as defined in \cite{ISO2001a-5349-1}. 
The initiation of the activity is marked by the creation of an \textit{IfcEvent}. Upon attaining the critical exposure action value (EAV) of $2.5 \,\text{m/s}^2$, a triggering mechanism schedules an \textit{IfcTask}, thereby assigning a safety instructor to intervene and mitigate the vibration-induced workload on the affected worker. Following the termination of the activity, another \textit{IfcEvent} containing the accumulated daily vibration exposure is created, enabling the precise tracking and verification of compliance with ISO 5349-1 limits.
Sensor-recognized activities are written into the twin as Activity Instances that can reuse the project’s schedule vocabulary (labels, scopes) when present, but remain valid even when no schedule file, look-ahead, or CPM alignment exists. This mapping exploits IFC’s extensible structure, where sensor activities are instantiated as new or augmented entities with attributes for safety metrics (e.g. $a_{hv}$ values and risk flags), ensuring interoperability without altering the core BIM model.

\section{Experiments and Results}

\subsection{SG-ICP}
\begin{figure}
    \centering
    \begin{subfigure}{0.48\linewidth}
        \centering
        \includegraphics[width=\linewidth]{"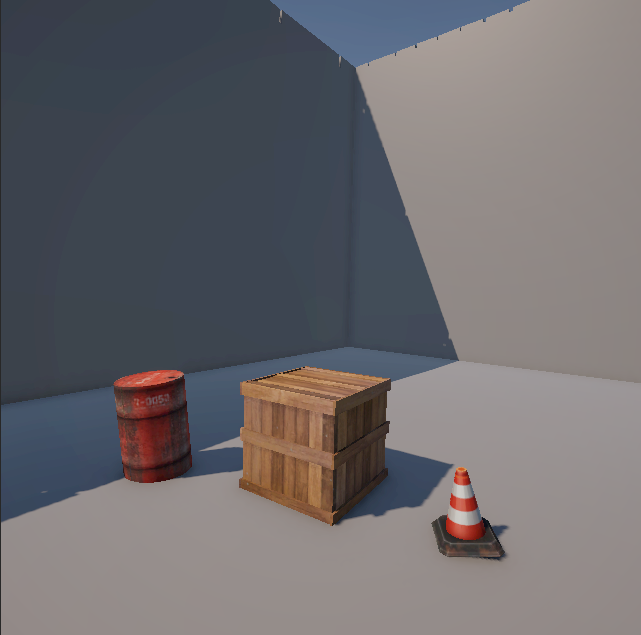"}
        \caption{Scenario 1}
        \label{fig:scen1}
    \end{subfigure}\hfill
    \begin{subfigure}{0.48\linewidth}
        \centering
        \includegraphics[width=\linewidth]{"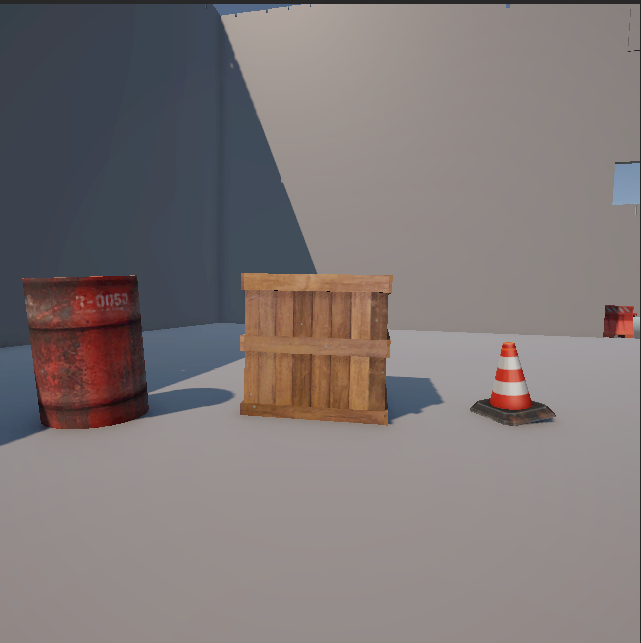"}
        \caption{Scenario 2}
        \label{fig:scen2}
    \end{subfigure}\hfill
    \begin{subfigure}{0.48\linewidth}
        \centering
        \includegraphics[width=\linewidth]{"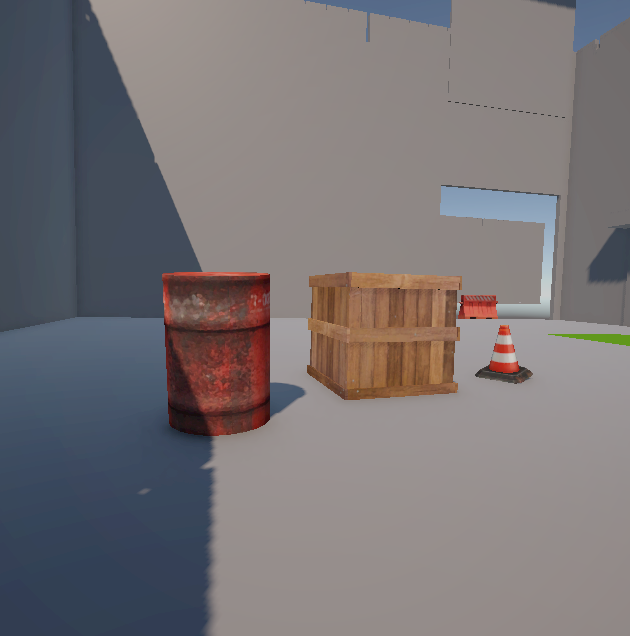"}
        \caption{Scenario 3}
        \label{fig:scen3}
    \end{subfigure}
        \begin{subfigure}{0.49\linewidth}
        \centering
        \includegraphics[width=\linewidth]{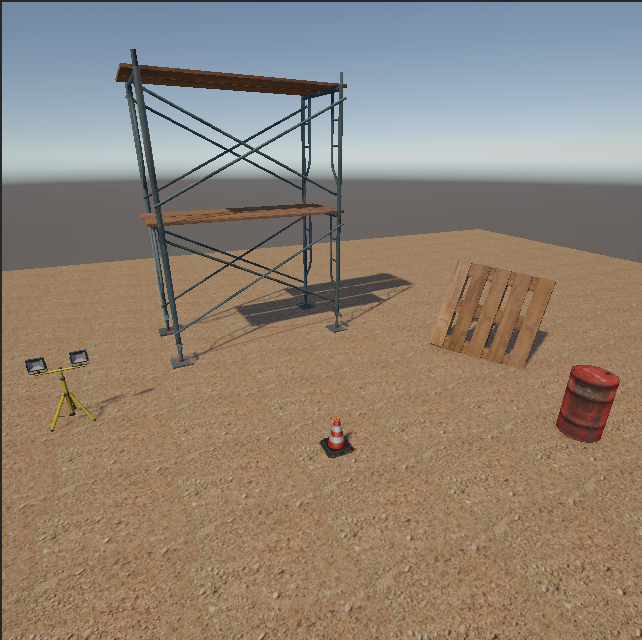}
        \caption{Scenario 4}
        \label{fig:scen4}
    \end{subfigure}
    \caption{Four evaluation scenarios used for the mesh-overlap analysis.}
    \label{fig:scenarios}
\end{figure}

Here we give mesh results on how well the meshes overlap given the ground truth and SG-ICP. We experiment in four different scenarios. One is the same setup with three different viewpoints and the fourth one is a variety of BIM families which would have poor feature description (scaffolding, tripod, wooden pallets, etc.). We experiment with 3 different perspectives to show the robustness under different feature sets. Both ICP implementations use Huber weights, restart poses, and randomize initial conditions to further improve convergence. The three scenarios can be seen in Figure \ref{fig:scenarios}. 

\begin{table}
\centering
\small
\setlength{\tabcolsep}{6pt}
\begin{tabular}{lccc}
\toprule
Metric & ICP & SG-ICP (proposed) & Improvement \\
\midrule
\multicolumn{4}{c}{Scenario 1} \\
\midrule
RMSE & 0.068321 & 0.007987 & 0.060334 (88.3\%~$\downarrow$) \\
Mean & 0.059600 & 0.005504 & 0.054096 (90.8\%~$\downarrow$) \\
Max  & 0.143498 & 0.026692 & 0.116806 (81.4\%~$\downarrow$) \\
\midrule
\multicolumn{4}{c}{Scenario 2} \\
\midrule
RMSE & 0.038275 & 0.034555 & 0.003720 (9.7\%~$\downarrow$) \\
Mean & 0.031619 & 0.028508 & 0.003111 (9.8\%~$\downarrow$) \\
Max  & 0.095449 & 0.072307 & 0.023142 (24.2\%~$\downarrow$) \\
\midrule
\multicolumn{4}{c}{Scenario 3} \\
\midrule
RMSE & 0.048497 & 0.017305 & 0.031192 (64.3\%~$\downarrow$) \\
Mean & 0.032258 & 0.014184 & 0.018074 (56.0\%~$\downarrow$) \\
Max  & 0.149865 & 0.041335 & 0.108530 (72.4\%~$\downarrow$) \\
\midrule
\multicolumn{4}{c}{Scenario 4} \\
\midrule
RMSE & 0.010916 & 0.011134 & -0.000218 (2.0\%~$\uparrow$) \\
Mean & 0.009282 & 0.009077 & 0.000205 (2.2\%~$\downarrow$) \\
Max  & 0.025163 & 0.033026 & -0.007863 (31.2\%~$\uparrow$) \\
\bottomrule
\end{tabular}
\caption{Hausdorff distances.}
\label{tab:icp_norm_improve_flip}
\end{table}

These scenarios represent varying levels of object occlusion and viewpoint challenges commonly encountered in construction environments, such as partial visibility due to site clutter or robot positioning constraints. The diversity in feature sets tests the algorithm's ability to handle sparse or repetitive geometries, which are prevalent in building sites with similar structural elements.

We quantify alignment quality using the Hausdorff distance to measure the alignment of the meshes. 
Quantitative metrics can be seen in Table \ref{tab:icp_norm_improve_flip}.

The results indicate that SG-ICP consistently outperforms standard ICP across all metrics, with particularly substantial improvements in Scenarios 1 and 3 (e.g., 88.3\% and 64.3\% RMSE reductions, respectively), where semantic-gravity biases help mitigate convergence to local minima caused by tilted orientations or limited features. This signals far fewer gross misalignments or outliers that typically drive false "deviation" flags in scan-to-BIM workflows. In contrast, the modest gains in S2 (9.7\% RMSE reduction) suggest scenarios with stronger initial alignments benefit less from the added priors, highlighting SG-ICP's value in challenging, real-world construction scans. This enhanced accuracy supports more reliable digital twin updates, potentially reducing errors in progress monitoring and robot path planning by ensuring upright and precise object placements.
\par
Even though SG-ICP received worse Hausdorff scores in Scenario 4, it has more reasonable semantic alignment as opposed to the unbiased ICP. We can see that the tripod is flipped upside down in the original ICP, converging to a local minimum. However, due to the upright bias, SG-ICP achieves better semantic alignment, which is what is materialized in the real-world. 

\subsection{Robot Exploration and BIM Alignment}
Here, we experiment by determining objects of interest (OOI) that need to be updated within a BIM that a robot would need to detect and align.

\begin{figure}
  \centering
  \begin{subfigure}{0.5\linewidth}
    \centering
    \includegraphics[width=\linewidth]{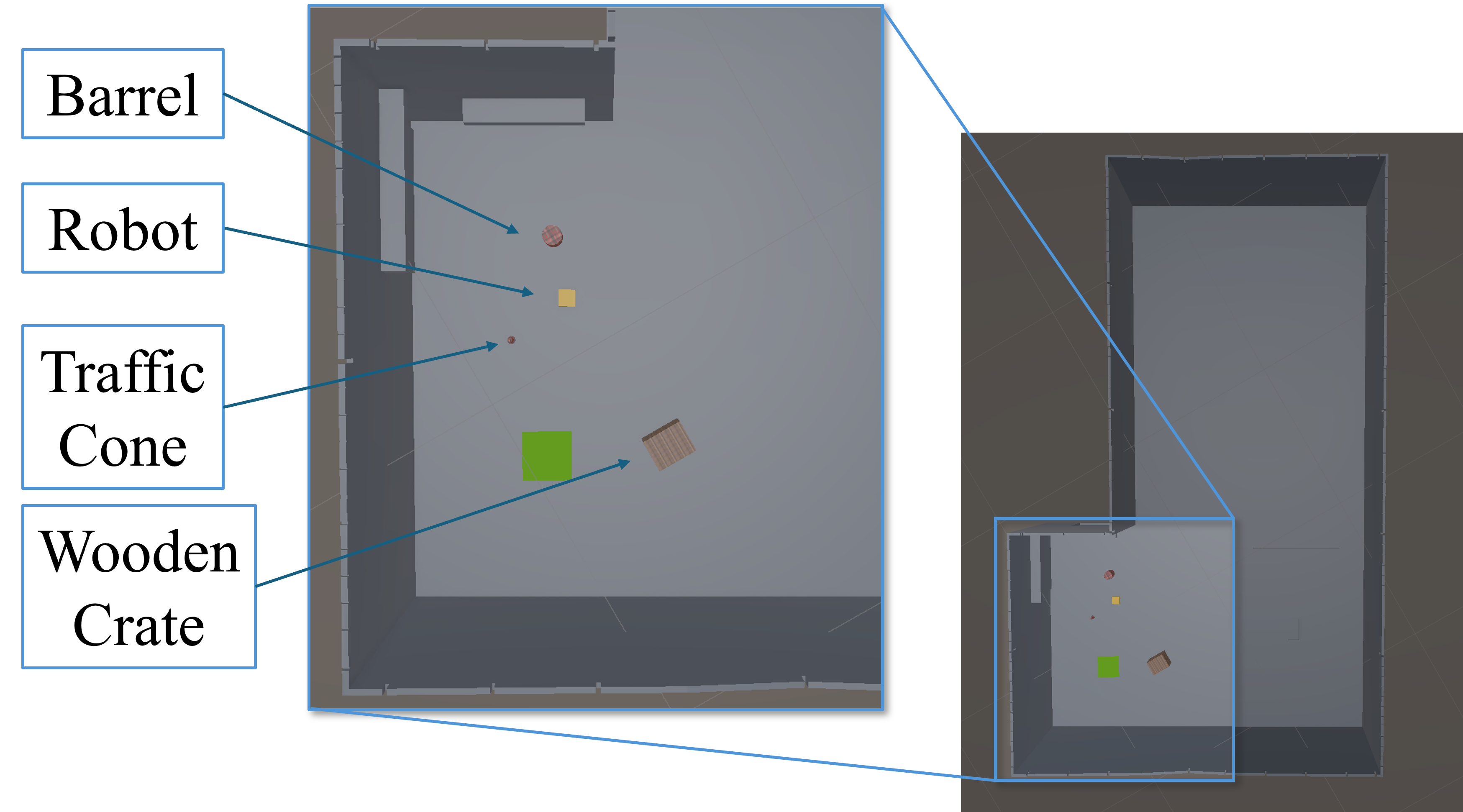}
    \caption{Real-world representation}
    \label{fig:Real_zoom}
  \end{subfigure}\hfill
  \begin{subfigure}{0.45\linewidth}
    \centering
    \includegraphics[width=\linewidth]{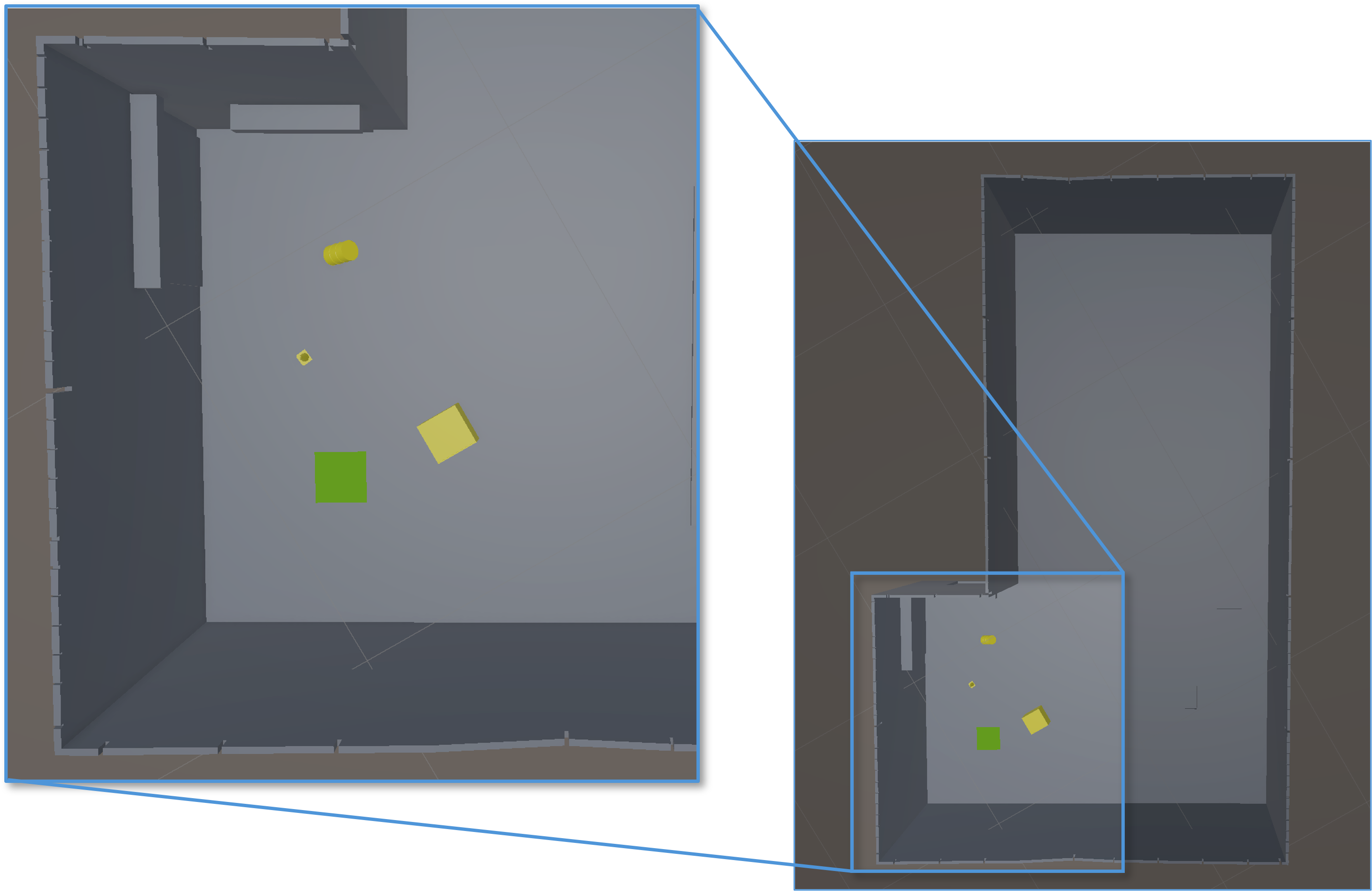}
    \caption{Digital Twin representation}
    \label{fig:BIM_zoom}
  \end{subfigure}
  \caption{Comparison of real-world scene and digital twin representation after ICP alignment}
  \label{fig:Real_and_BIM_comparison}
\end{figure}

The OOIs under consideration in this study include traffic cone, barrel, and wooden crate. As demonstrated in Figure \ref{fig:Real_and_BIM_comparison}, the ICP aligns the objects in the Digital Twin with the ground truth given in the real-world scenario. 

This cone-barrel-crate study demonstrates end-to-end behavior: the alignment showcases the framework's capability to integrate robot-collected data into BIM models, enabling automated detection of temporary site elements like safety barriers or storage items that may deviate from as-designed plans. Such functionality is significant for construction robotics applications, as it facilitates real-time deviation identification, enabling immediate re-planning around temporary obstructions and improving site safety and efficiency by allowing robots to adapt to dynamic environments without manual intervention.

\begin{figure}
  \centering
  \begin{subfigure}{0.48\linewidth}
    \centering
    \includegraphics[width=\linewidth]{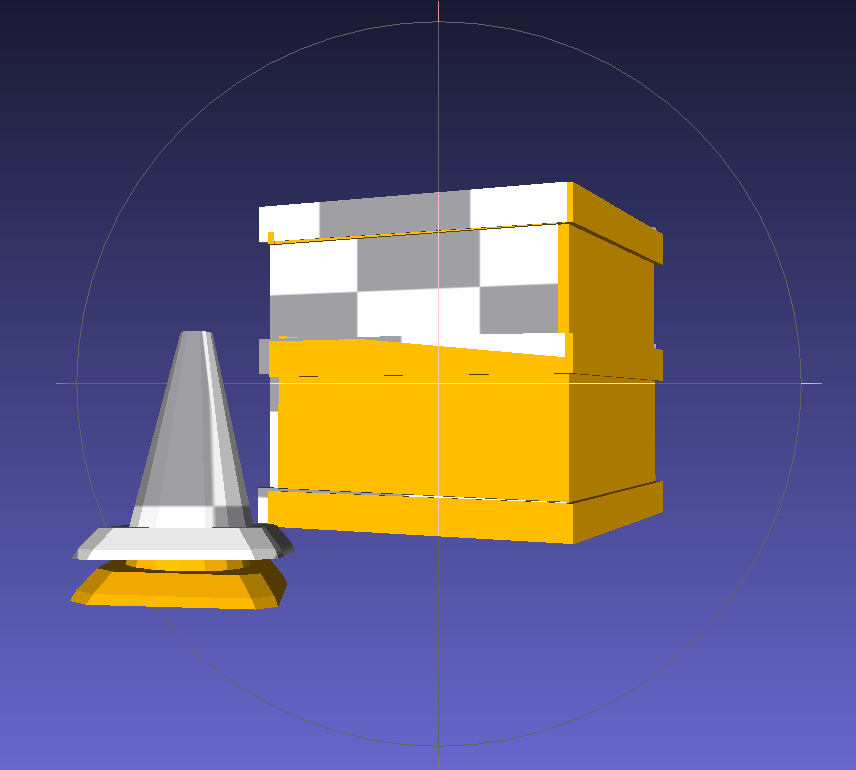}
    \caption{SG-ICP}
    \label{fig:bias_biased_1}
  \end{subfigure}\hfill
  \begin{subfigure}{0.48\linewidth}
    \centering
    \includegraphics[width=\linewidth]{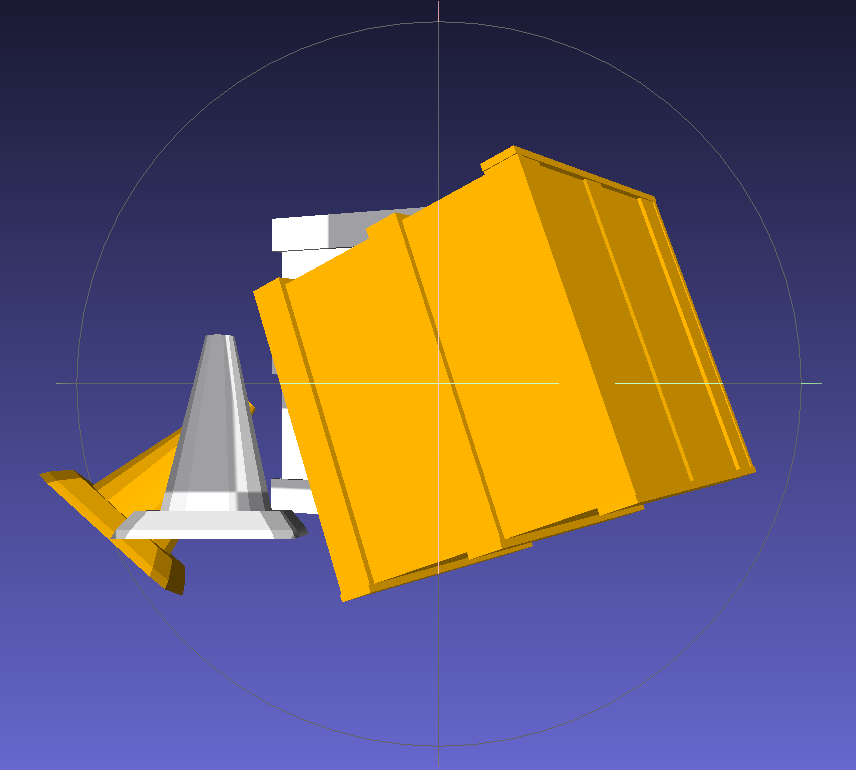}
    \caption{ICP}
    \label{fig:bias_unbiased_1}
  \end{subfigure}
  \caption{SG-ICP vs ICP in Scenario 1}
  \label{fig:bias_results_S1}
\end{figure}

\begin{figure}
  \centering
  \begin{subfigure}{0.48\linewidth}
    \centering
    \includegraphics[width=\linewidth]{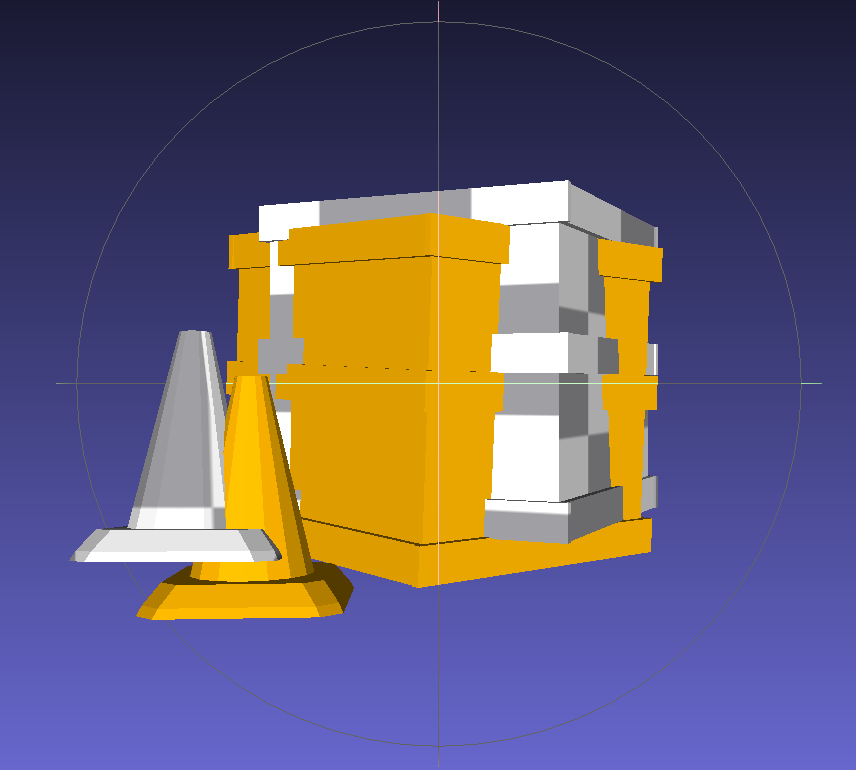}
    \caption{SG-ICP}
    \label{fig:bias_biased_2}
  \end{subfigure}\hfill
  \begin{subfigure}{0.48\linewidth}
    \centering
    \includegraphics[width=\linewidth]{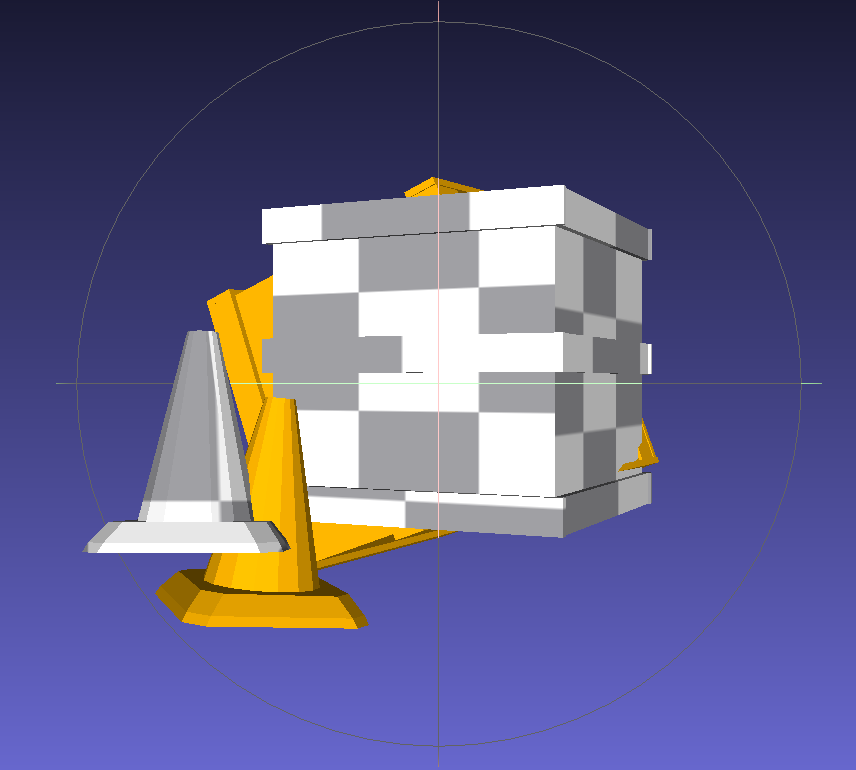}
    \caption{ICP}
    \label{fig:bias_unbiased_2}
  \end{subfigure}
  \caption{SG-ICP vs ICP in Scenario 2}
  \label{fig:bias_results_S2}
\end{figure}

\begin{figure}
  \centering
  \begin{subfigure}{0.48\linewidth}
    \centering
    \includegraphics[width=\linewidth]{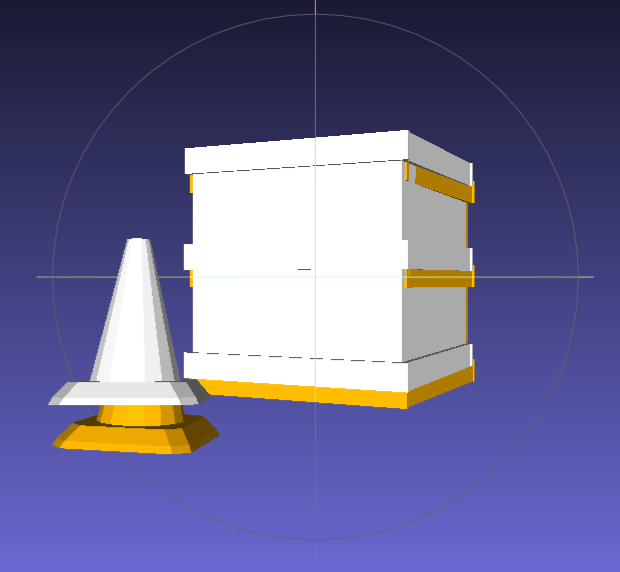}
    \caption{SG-ICP}
    \label{fig:bias_biased_3}
  \end{subfigure}\hfill
  \begin{subfigure}{0.48\linewidth}
    \centering
    \includegraphics[width=\linewidth]{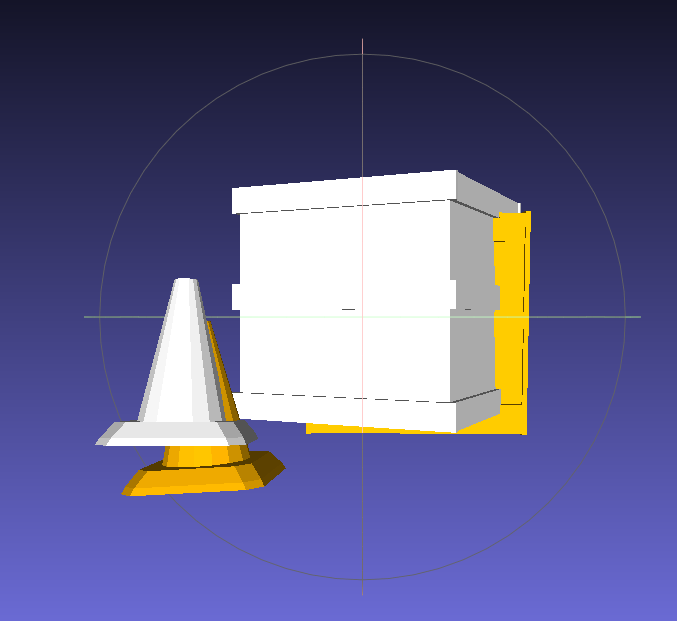}
    \caption{ICP}
    \label{fig:bias_unbiased_3}
  \end{subfigure}
  \caption{SG-ICP vs ICP in Scenario 3}
  \label{fig:bias_results_S3}
  \end{figure}

A more thorough examination shows the performance difference between ICP and SG-ICP, illustrating each scenario in Figures \ref{fig:bias_results_S1}, \ref{fig:bias_results_S2} and \ref{fig:bias_results_S3}.

Visually, SG-ICP maintains upright and accurate orientations in all scenarios, avoiding the misalignments and tilts evident in standard ICP outputs, which often result from local optima traps in feature-sparse views. Scenario 1 shows that the standard ICP algorithm converges to a physically implausible tilt, while SG-ICP preserves an upright solution and corrects residual yaw and pitch, producing a visibly crisper mesh overlap. In Scenario 2, both methods achieve comparable fits, and the SG-ICP advantage is small. This suggests that when features are plentiful and overlap is good, the added prior neither harms nor materially changes the solution, an important property for field robustness (the bias "gets out of the way" when not needed). Scenario 3 again shows SG-ICP resisting yaw drift and reducing boundary overhangs that inflate the Hausdorff maximum under ICP. Enforcing such physically plausible alignments is beneficial for construction robotics applications where incorrect poses could lead to navigation errors or false deviation alerts in digital twins.

\begin{figure}
  \centering
  \begin{subfigure}{0.48\linewidth}
    \centering
    \includegraphics[width=\linewidth]{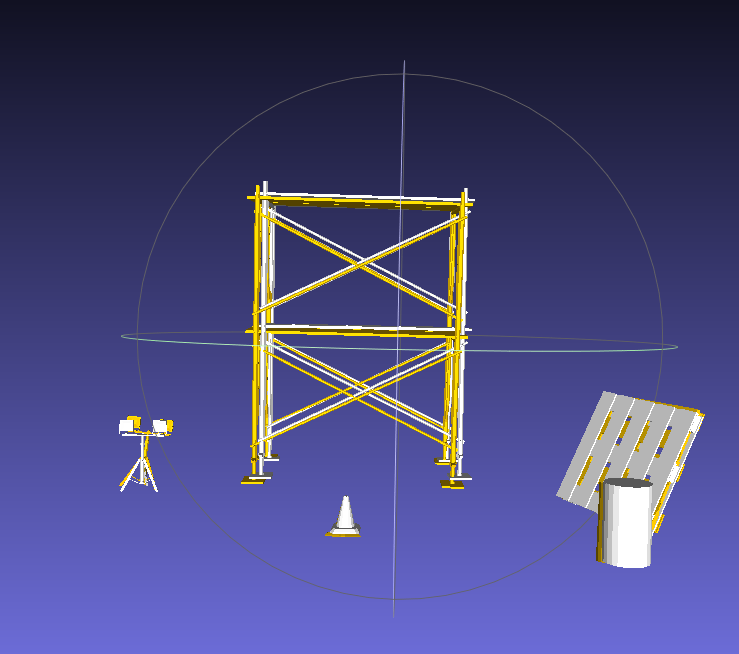}
    \caption{SG-ICP}
    \label{fig:bias_biased_4}
  \end{subfigure}\hfill
  \begin{subfigure}{0.48\linewidth}
    \centering
    \includegraphics[width=\linewidth]{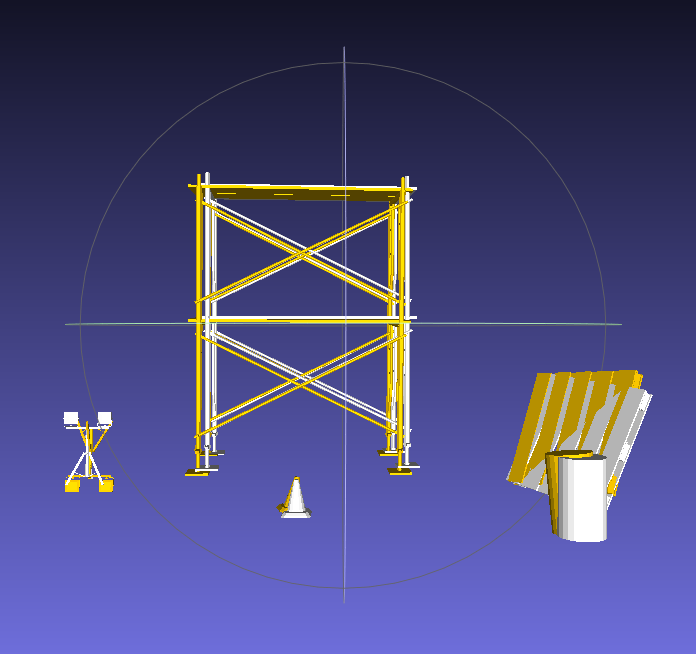}
    \caption{ICP}
    \label{fig:bias_unbiased_4}
  \end{subfigure}
  \caption{SG-ICP vs ICP in Scenario 4}
  \label{fig:bias_results_S4}
\end{figure}

\subsection{Hand-Arm Vibration Activity Mapped in the DT}

\begin{figure}
  \centering
  \begin{subfigure}{0.48\linewidth}
    \centering
    \includegraphics[width=\linewidth]{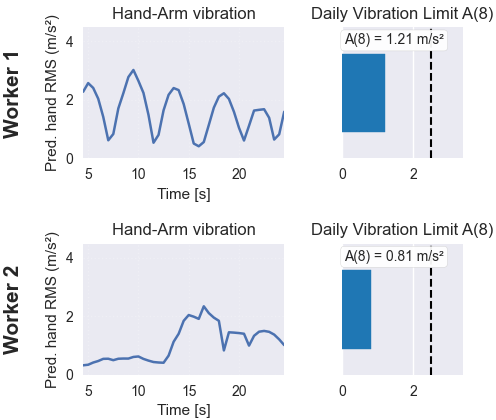}
    \caption{Time t\_1}
    \label{fig:HAV_dashboard_timestep_1}
  \end{subfigure}\hfill
  \begin{subfigure}{0.48\linewidth}
    \centering
    \includegraphics[width=\linewidth]{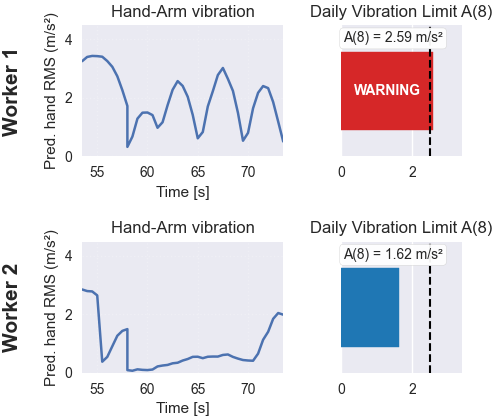}
    \caption{Time t\_2}
    \label{fig:HAV_dashboard_timestep_2}
  \end{subfigure}
  \caption{Integration of HAV monitoring into the updated digital twin, with automatic warning at the daily vibration limit.}
  \label{fig:HAV_dashboard}
\end{figure}

As illustrated in Figure \ref{fig:HAV_dashboard}, the HAV dashboard systematically records and processes vibration data, subsequently generating visual warnings upon violation of the EAV. Simultaneously, the \textit{IfcEvent} start, end, accumulated daily vibration and the subsequent \textit{IfcTask} of intervention are written to the IFC file.

\section{Discussion}

\subsection{Implications of Geometric Alignment Results}

The experimental outcomes for SG-ICP highlight its superior performance in aligning robot-captured meshes with BIM models, as evidenced by the substantial reductions in Hausdorff distances across diverse viewpoints. Notably, the high improvements in Scenarios 1 and 3 demonstrate how semantic-gravity integration addresses common ICP pitfalls, such as sensitivity to initial poses in environments with ambiguous features, analogous to construction sites with often occluded elements like scaffolding or equipment stacks. This robustness enhances the reliability of digital twins for as-built verification, potentially minimizing costly rework by enabling early detection of geometric deviations.

Furthermore, the successful alignment of OOIs in the robot exploration experiments, as shown in Figure \ref{fig:Real_and_BIM_comparison}, illustrates SG-ICP's practical value for autonomous site monitoring. By ensuring physically consistent orientations (e.g., upright objects resistant to tipping), the method supports safer robot navigation and more accurate progress tracking, laying a foundation for integrating semantic updates like activity recognition in dynamic construction workflows.

Finally, the mixed magnitude of improvements across scenarios argues for adaptive priors. Specifically, schedule and scene context (when available) or learned scene cues can modulate the semantic-gravity weight, preserving the strong performance in difficult cases without over-constraining easy ones. This aligns with our methodology goal of agentic parameterization-LLM-mediated "common sense" informs the bias where helpful and abstains where unnecessary.

\subsection{Practical Implementation of BIM2RDT in Construction Settings}

The successful deployment of BIM2RDT within active construction projects requires careful consideration of how this methodology integrates with established project delivery processes, existing technology infrastructure, and field operations management. Construction practitioners face unique challenges when implementing advanced digital twin technologies that extend beyond pure technical functionality to encompass organizational change management, workflow coordination, and economic feasibility within competitive project environments.

\subsubsection{Preconstruction and Construction Execution Workflows}

During preconstruction phases, the methodology requires integration with existing BIM coordination workflows where architects, engineers, and major trade contractors perform clash detection and constructability reviews. The semantic object detection components need to be configured during this phase by identifying which BIM families will require regular geometric verification and safety monitoring during construction execution.

This preconstruction setup process involves collaboration between the BIM coordination team and safety management personnel to establish the activity entity classifications that will be monitored throughout the project. Rather than creating entirely new safety documentation procedures, the system builds upon existing safety planning processes by extending them to include sensor-based activity recognition protocols. Safety managers can incorporate the wearable vibration monitoring requirements into existing personal protective equipment protocols, treating the sensors as standard safety equipment rather than specialized research instruments.

The construction phase implementation requires coordination between field superintendents, safety supervisors, and technology coordinators to manage the daily operations of both robotic scanning activities and worker-worn sensor monitoring. Superintendents need clear protocols for scheduling robot deployment activities that complement rather than interfere with active construction operations. The robot scanning process should integrate with existing quality control inspections and progress documentation activities, allowing field teams to accomplish multiple objectives during single site traversals.

\subsubsection{Technology Infrastructure and Existing Systems Integration}

The BIM2RDT implementation must demonstrate clear integration pathways with established software platforms rather than requiring complete technology stack replacements. The IFC-based activity entity approach facilitates this integration by ensuring that detected safety events and geometric updates can be imported into existing project management software through standard BIM data exchange protocols.
The implementation strategy should minimize the number of additional software interfaces that field personnel must learn while maximizing the value delivered through enhanced project visibility and safety monitoring capabilities.  Some specific considerations regarding model, control, and scope include:
\begin{itemize}
\item Network and time synchronization. The event bus can run over site Wi-Fi or a private LTE hotspot. Time-sync wearables, robot, and tablets to a shared source (PTP or NTP) so activity windows and poses line up without manual massaging.
\item BIM readiness. The robot uses BIM geometry and semantics as the \textit{a priori} map. Thus, the implementation calls for a quick model conditioning pass in terms of details such as level of development (LoD): purge stray elements, classify major families (walls, columns, cores, permanent equipment), and tag off-limits areas. Per-element millimeter fidelity isn’t required; consistent categories and true elevations matter more than ornamental detail.
\item Project coordinates. It is important to tie the BIM to the survey control network (project base point + survey point) and publish a single “site frame” for all devices. Using check shots at each level (stair landings, core corners) so the robot’s localization can be sanity-checked by field engineers is encouraged.
\end{itemize}

\subsubsection{Data governance and worker acceptance}

Data management protocols require clear definition of responsibilities between general contractors, subcontractors, and specialty technology providers. The geometric updates generated through SG-ICP processing and safety events detected through wearable monitoring create valuable project documentation that multiple stakeholders need to access and utilize. Construction practitioners require clear data ownership agreements and coordination protocols that specify which parties are responsible for data collection, processing, validation, and long-term archival. In addition, the following should be given serious consideration:

\begin{itemize}

\item Privacy \& consent. Wearables should be opt-in with clear signage. Default to pseudonymized IDs in the UI; reveal identity only to authorized safety staff. Store raw sensor windows behind access controls; surface interpreted metrics (e.g., awRMS, dose) in daily reports.
\item System of record. Treat the twin as a coordination aid, not a contractual as-built, unless your QA/QC team promotes selected snapshots through formal verification.
\end{itemize}

\subsubsection{Field Personnel Training and Adoption Strategies}

The success of BIM2RDT implementation depends heavily on acceptance and proper utilization by field personnel who may have varying levels of comfort with advanced technology systems. Construction foremen and craft workers need straightforward protocols for managing wearable safety sensors that fit naturally into existing personal protective equipment routines without creating additional administrative burdens or work interruptions. 
Training programs should emphasize practical benefits that directly improve daily work experiences rather than focusing primarily on technical system capabilities. When field personnel understand that the vibration monitoring system provides real-time feedback that helps them avoid potential health risks while maintaining productivity, adoption becomes much more natural than when the system is presented primarily as a compliance monitoring tool.

\subsubsection{Economic Considerations and Return on Investment}
The implementation cost analysis must account for both direct technology expenses including robot hardware, sensor systems, and software licensing, and indirect costs including training time, workflow modifications, and potential productivity impacts during initial implementation phases. 
The return on investment calculations should emphasize quantifiable benefits that align with established construction project success metrics. Reduced rework costs through improved geometric accuracy verification provide direct cost savings that can be measured against existing quality control expenses. Enhanced safety monitoring capabilities reduce potential liability exposure and worker compensation claims while demonstrating compliance with evolving occupational health regulations.

\section{Conclusion}

This paper introduced BIM2RDT, an agentic AI framework that transforms static Building Information Models into dynamic, robot-ready digital twins with an emphasis on safety. By integrating BIM data, robot-collected point clouds, and IoT sensor streams for activity monitoring, the framework provides a robust solution for real-time site understanding. The proposed Semantic-Gravity ICP (SG-ICP) algorithm, which leverages LLM-based reasoning to inform physical priors, was shown to significantly improve point cloud registration accuracy, achieving RMSE reductions of up to 88.3\% over standard methods and ensuring physically plausible alignments. Furthermore, the integration of HAV monitoring demonstrated the framework’s capability to map safety-critical events to the digital twin, enabling proactive interventions.

\subsection{Limitations}

Despite its contributions, the work has limitations. First, the experiments were conducted in controlled settings, and the framework's scalability to large, highly dynamic construction sites with diverse and unanticipated objects remains to be validated. Second, the framework’s dependence on high-quality BIM models as a priori maps may limit applicability in projects with incomplete or poorly structured digital models. Third, while integrating multiple data streams, the framework does not yet fully address the complexities of real-time data synchronization and conflict resolution, especially for very large-scale projects.

\subsection{Future Work}

Future work will focus on conducting extensive field trials on active construction sites to validate robustness, developing error-correction mechanisms for imperfect BIM inputs, and optimizing algorithms for edge computing to improve real-time performance and scalability. Research should also focus on expanding the agentic AI orchestrator to handle broader sensor fusion scenarios (extending the same agentic layer that already assigns semantic priors for SG-ICP). For example, for the HAV detection scenario presented here as a proof-of-concept, after vibration metric computation (e.g., frequency-weighted RMS acceleration $a_{hv}$ compared to ISO thresholds), an agent can process IoT data from wearables to classify activities, inferring labels like "prolonged HAV exposure" based on signal patterns and contextual cues from robot observations. A semantic mapping agent can then cross-reference these with BIM elements and using LLM-based reasoning, aligns detected activities to \textit{IfcTask} instances. 

\section{Acknowledgments}

The presented work has been supported by the U.S. National Science Foundation (NSF) CAREER Award through grant No. 2047138. Any opinions, findings, conclusions, and recommendations expressed in this paper are those of the authors and do not necessarily represent those of the NSF.

\bibliography{ascexmpl-new}
\vspace{12pt}

\end{document}